\documentclass[journal]{IEEEtran}

\ifCLASSINFOpdf
\else
   \usepackage[dvips]{graphicx}
\fi
\usepackage{url}

\usepackage{graphicx}
\usepackage{tikz} 
\usepackage{amsmath,amsfonts}
\usepackage{hyperref}
\usepackage[numbers,sort&compress]{natbib}
\author{
  \IEEEauthorblockN{Abhilash Jain}
  \IEEEauthorblockA{
    \textit{Aalto University}\\
  abhilash.jain@aalto.fi\\}
    \IEEEauthorblockN{Aku Ruohe}
  \IEEEauthorblockA{
    \textit{Aalto University}\\
  aku.ruohe@aalto.fi\\}
  \IEEEauthorblockN{Stig-Arne Grönroos }
  \IEEEauthorblockA{
    \textit{Aalto University}\\
  stig-arne.gronroos@aalto.fi\\}
  \IEEEauthorblockN{Mikko Kurimo}
  \IEEEauthorblockA{
    \textit{Aalto University}\\
  mikko.kurimo@aalto.fi\\}
  
}

\begin{document}

\pgfdeclarelayer{bg}
\pgfdeclarelayer{foreground}
\pgfsetlayers{bg,main,foreground}

\title{Finnish Language Modeling with Deep Transformer Models}

\maketitle
\begin{abstract}
Transformers have recently taken the centre stage in language modeling after LSTM's were considered the dominant model architecture for a long time. In this project, we investigate the performance of the Transformer architectures-BERT and Transformer-XL for the language modeling task. We use a sub-word model setting with the Finnish language and compare it to the previous State of the art (SOTA) LSTM model. BERT achieves a pseudo-perplexity score of 14.5, which is a first such measure achieved as far as we know. Transformer-XL improves upon the perplexity score to 73.58 which is 27\% better than the LSTM model.

\end{abstract}

\section{Introduction}
\label{sec:introduction}
Language modeling is a probabilistic description of language phenomenon. It provides essential context to distinguish words which sound similar and therefore has one of the most useful applications in Natural Language Processing (NLP) especially in downstreaming tasks like Automatic Speech Recognition (ASR). Recurrent Neural Networks (RNN) especially Long Short Term Memory (LSTM) networks \cite{hochreiter1997long}  have been the typical solution to language modeling which do achieve strong results. In spite of  these results, their fundamental sequential computation constraint has restricted their use in the modeling of long-term dependencies in sequential data. To address these issues Transformer architecture was introduced. Transformers relies completely on an attention mechanism to form global dependencies between input and output. It also offers more parallelization and has achieved SOTA results in language modeling outperforming LSTM models \cite{NIPS2017_7181}.

In recent years,we have seen a lot of development based on this standard transformer models particularly on unsupervised pre-training(\cite{Radford2018ImprovingLU,DBLP:journals/corr/abs-1810-04805,Dai2019TransformerXLAL,yang2019xlnet,DBLP:journals/corr/abs-1802-05365,DBLP:journals/corr/abs-1801-06146} which have set state-of-the art results on multiple NLP benchmarks. One such model architecture has been the Bidirectional Encoder Representations from Transformers (BERT) model which uses a deep bidirectional transformer architecture.

Another architecture of interest would be the Transformer-XL, which introduces the notion of recurrence in a self-attention model.

The primary research focus though has been mostly on English language for which abundant data is present. It is interesting to see the performance of these models for an agglutinative language like Finnish, which is morphologically richer than English.

In this project, we explore the implementation of Transformer-based models (BERT and Transformer-XL) in language modeling for Finnish. We will use the same training data as in \cite{Aaltodoc:http://urn.fi/URN:ISBN:978-952-60-8566-1} so that we can do fair comparisons with the performance of the LSTM models.
Also, as the BERT model is a bi-directional transformer, we will have to approximate the conditional probabilities given a sequence of words. We also experiment with using sub-word units with Transformer-XL to cope with the large vocabulary problems associated with the Finnish Language. With smaller units, the modeled sequences are longer, and we hope that the recursive XL architecture can allow us to still model long term effects. To the best of our knowledge this is the first work with the Finnish language to use the following:
\begin{itemize}
    \item Approximation of perplexity using a BERT architecture
    \item Using Transformer-XL architecture with sub-word units.
    \item Comparison of Transformer and LSTM models as language models in the same comparable settings with an agglutinative language.
\end{itemize}

\section{Background \& Methods}
The goal of an language model is to assign meaningful probabilities to a sequence of words. 
Given a set of tokens $\mathbf{X}=(x_1,....,x_T)$, where $T$ is the length of a sequence, our task is to estimate the joint conditional probability $P(\mathbf{X})$ which is \begin{equation}
    \label{cond}
    P(\mathbf{X})=\prod_{i=1}^{T} p\left(x_{i} | x_{1}, \ldots, x_{i-1}\right) ,
\end{equation} were $(x_{1}, \ldots, x_{i-1})$ is the context. An Intrinsic evaluation of the performance of Language Models is perplexity (PPL) which is defined as the inverse probability of the set of the tokens and taking the $T^{th}$ root were $T$ is the number of tokens
 \begin{equation}
    \label{ppl}
    PPL(\mathbf{X})= P(\mathbf{X})^{-1/T}.
\end{equation}
In our two approaches we use transformer based architectures: BERT and Transformer-XL as mentioned before. Calculating the auto-regressive $P(\mathbf{X})$ for the transformer-XL is quite straight-forward as the model is unidirectional but it doesn't factorize the same way for a bi-directional model like BERT.

BERT's bi-directional context poses a problem for us to calculate an auto-regressive joint probability. A simple fix could be that we mask all the tokens $\mathbf{x}_{>i}$ and we calculate the conditional factors as we do for an unidirectional model. By doing so though, we loose upon the advantage of bi-directional context the BERT model enables. We propose an approximation of the joint probability as,

\begin{equation}
\label{approx}
    P(\mathbf{X}) \approx \prod_{i=1}^{T} p\left(x_{i} | x_{1}, \ldots, x_{i-1}, x_{i+1}, \ldots, x_{T}\right).
\end{equation}
This type of approximations has been previously explored with Bi-directional RNN LM's \cite{inproceedings} but not for deep transformer models. We therefore, define a pseudo-perplexity score from the above approximated joint probability.

The original BERT has two training objectives: 'Masked language modelling', in which you mask input tokens randomly and then predict the masked tokens using the left and right context. Additionally, there is the 'next sentence prediction' task that jointly trains text-pair representations. For training the Masked language model the original BERT used Byte Pair Encoding (BPE) \cite{10.5555/177910.177914} for subword tokenization \cite{DBLP:journals/corr/SennrichHB15}.For example the rare word "unaffable" to be split up into more frequent subwords such as ["un", "\text{\#\#aff"}, \text{"\#\#able"}]. To remain consistent with experiments performed with LSTM's we use the morfessor for the subword tokenization in the Finnish Language. In Addition, we also apply boundary markers as in (Table \ref{Tab:markings}) and train two separate models using this distinction. We train with left-marked markings as the original BERT was trained with such a scheme and the left+right-marked as it was the previous SOTA with the Finnish Language. For the transformer-XL experiments, we just train with the left+right marked scheme.

\begin{table}[h!]
    \footnotesize
    \centering
    \caption {Two methods of marking subword units such that the original sentence 'two slippers' is reconstructed}
    \begin{tabular}{l|  l  }
        subword marking & Example \\
        \hline
        left+right-marked (+m+) & two slipp+ +er+ +s \\
        left-marked (+m) & two slipp +er +s \\
        \hline
    \end{tabular}
    \label{Tab:markings}
\end{table}

\begin{figure*}[t]
    \centering
    \includegraphics[scale=0.9]{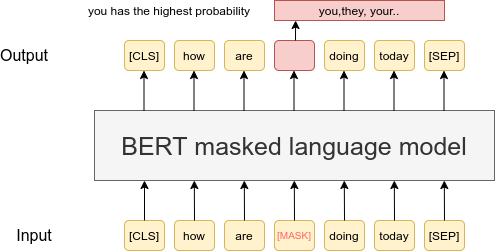}
    \caption{BERT-Original sentence 'how are you doing today'}
    \label{fig:BERT_label}
\end{figure*}

The Next Sentence Prediction (NSP) is a binary classification task which predicts whether two segments follow each other in the original text. This pre-training task was proposed to further improve the performance on downstreaming tasks, like Natural Language Inference(NLI) but in reality removing the NSP loss matches or slightly improves the downstream task performance \cite{DBLP:journals/corr/abs-1907-11692}. In this paper, we have omitted the NSP task from the BERT pre-training procedure and changed the input from a SEGMENT-PAIR input to a SINGLE SEGMENT input. As seen in (Fig \ref{fig:BERT_label})

Transformer-XL introduced the notion of recurrence in self-attention by caching the hidden state sequence to compute the hidden states of a new segment. It also introduces a novel relative positional embedding scheme and both of them combined address the issue of fixed context lengths.
Transformer-XL as mentioned is a unidirectional deep transformer architecture, therefore the perplexity can be calculated as (Eq \ref{ppl}). The only change is in the input format, were we use sub-word units rather than whole word units as Finnish is morphologically richer than English.

\section{Data}
\label{sec:data}

The Finnish text data used for the language modeling task is provided by \cite{ftc-korp_en}. The dataset consists mainly of newspapers and books of around 144 million word tokens and 4.2 million unique tokens.
We use a Morfessor 2.0 \cite{smit2014morfessor} using the basic unsupervised Morfessor Baseline algorithm \cite{10.1145/1187415.1187418} with a corpus weight parameter ($\alpha$) of 0.001. We have a vocabulary of 34K subword tokens for the left+right-marked (+m+) markings and 19K subword tokens for the left-marked (+m) markings. We also pre-process the data to remove any punctuation marks such that we can use the same data with an ASR system. The input is one sentence per line and we shuffle the sentences at each epoch. The data is randomly divided into- training dataset and a validation dataset. The test dataset consists of 2850 Finnish news articles obtained from the Finnish national broadcaster YLE.

\section{Experiments \& Results}
\subsection{BERT}
All BERT experiments were trained for 500K steps. The code was written in Python and we used the Tensorflow libraries to create the models. The experiments were trained on a single NVIDIA Tesla V100 32 GB graphic card. The data was first processed into Tensorflow records as the input to the model. The set of hyperparameters which we found optimal after experimenting with different sets are in  (Table \ref{Tab:BERT}).

\begin{table}[h!]
    \footnotesize
    \centering
    \caption {BERT hyperparameters}
    \begin{tabular}{l l  }
        \hline
        Number of hidden layers & 20 \\
        Hidden size of transformer & 896 \\
        Number of attention heads & 16 \\
        Intermediate size(Size of the feed forward layer) & 3584 \\
        hidden activation function & Gaussian Error Linear Units \\
        dropout probability & 0.1 \\
        max position embeddings & 300 \\
        \hline
    \end{tabular}
    \label{Tab:BERT}
\end{table}
This set of parameters were chosen as there training performances were better than smaller models on modelling the long sequences of sub-words. We use the Adam optimizer \cite{Kingma2014AdamAM} same as the English BERT. A maximum sequence length of 300 encompasses 98 percent of the training data and also allows us to fit larger models on the GPU card. Hyper-parameter optimization is very difficult in case of these models as they take around 15 days to train given the resources. The hyperparameter choices were therefore more dependant on the original BERT with little tweaks. We assess the training performance of the the model in the (Table \ref{Tab:BERT-Train}).
\begin{table}[h!]
    \footnotesize
    \centering
    \caption {BERT training performance}
    \begin{tabular}{l | l | l }
        Model & Masked LM Loss & Masked LM Accuracy \\
        \hline
        left+right-marked (+m+) &  2.24 & 0.56\\
        left-marked (+m) & 2.03 & 0.59 \\
        \hline
    \end{tabular}
    \label{Tab:BERT-Train}
\end{table}

When we train the BERT model we mask some percentage of the input tokens at random, and then predict those masked tokens, this is known as Masked LM. The masked LM loss, refers specifically to the loss when the masked language model predicts on the masked tokens. The masked LM accuracy refers specifically to the accuracy with which the model predicts on the masked tokens. The loss for both the models are far off from the Masked LM loss of the English BERT, key difference being the pre-training data for both the language models are quite different. Google training their model on 3.3 Billion words from BooksCorpus \cite{DBLP:journals/corr/ZhuKZSUTF15} and the English Wikipedia and our model being trained on 144 million words. Comparing the two Finnish models, the left-marked model has a better training performance than left+right-marked model.

The results of the pseudo-perplexity described in the previous section to evaluate the above models on the test data-set is in table (Table \ref{Tab:BERT-Test}).The test dataset is of a different context when compared to the training data, and interestingly both the models are quite confident when it comes to the test dataset. The pseudo-perplexity values of left-marked are lower when compared to left-right-marked signifying that it is more confident.
 
We cannot directly compare the perplexity scores BERT model with a unidirectional LSTM model as both are calculated in a different manner. We can experiment to compare it with a Bi-directional LSTM or use a downstreaming task to compare both the performances. We could also randomly mask tokens and then compare the prediction accuracy on the masked tokens.
\begin{table}[h!]
    \footnotesize
    \centering
    \caption {BERT Test performance}
    \begin{tabular}{l | l  }
        Model & Pseudo perplexity  \\
        \hline
        left+right-marked (+m+) & 17.1\\
        left-marked (+m) & 14.5\\
        \hline
    \end{tabular}
    \label{Tab:BERT-Test}
\end{table}

\subsection{Transformer-XL}
All Transformer-XL experiments are also trained equally for 500K steps. The code was written in Python and we used  the PyTorch libraries for model creation. The experiments were trained on a single NVIDIA Tesla V100 32 GB graphic card. Two sets of hyperparameters were chosen to be compared after some initial optimization and are in (Table \ref{Tab:trxl})
\begin{table}[h!]
    \footnotesize
    \centering
    \caption {Tr-XL hyperparameters}
    \begin{tabular}{l l | l}
        Hyperparameters & Model 1 & Model 2\\
        \hline
        Number of hidden layers & 4 & 4 \\
        Hidden size of transformer & 512 & 1024 \\
        Number of attention heads & 8 & 8  \\
        Size of attention head & 80 & 128 \\
        Intermediate size(Size of the feed forward layer) & 2048 & 4096 \\
        Warmup & 10000 & 40000 \\
        Batch-size & 64 & 224 \\
        Segment Length & 150 & 32 \\
        Memory Length & 150 & 32 \\
        \hline
    \end{tabular}
    \label{Tab:trxl}
\end{table}
From the above parameter choice, we wanted to experiment, whether providing more Segment and Memory length is advantageous (longer context) than a larger model. These parameters where chosen after some hyperparameter optimization. Same as for BERT we use the Adam optimizer, but we also use a cosine annealing learning rate scheduler to speed-up training \cite{DBLP:journals/corr/LoshchilovH16a}. The training performance results are in (Table \ref{Tab:trxl-train})

\begin{table}[h!]
    \footnotesize
    \centering
    \caption {Tr-XL training perplexity scores}
    \begin{tabular}{l | l | l }
        Model & Mem-seg len    \\
              & 150-150 & 32-32 \\ 
        \hline
        left+right-marked (+m+) & 45.22 & 33.86\\
        left-marked (+m) & 47.83 & 35.78 \\
        \hline
    \end{tabular}
    \label{Tab:trxl-train}
\end{table}
As opposed to BERT, the left+right-marked models have a better training performance than their counterpart. Interestingly the larger model trains much better when compared to providing larger contexts. The same set of parameters for the 32-32 model cannot be replicated for 150-150 model as the latter takes a lot of space on the GPU card. The test set is same as that used with BERT and the results are in (Table \ref{Tab:trxl-test}). The test performance is similar to that of the training performance with left-right-marked large model(32-32) performing the best.
We can directly compare the perplexity scores with the previous best \cite{Hmm-dmm} as both are unidirectional models, Transformer-XL model has outperformed the latter by 27\%. 
\begin{table}[h!]
    \footnotesize
    \centering
    \caption {Tr-XL test perplexity scores, (-): The experiment models are not available}
    \begin{tabular}{l | l | l | l }
        Model & Mem-seg len & &   \\
              & 150-150 & 32-32 & (prev best) \\ 
        \hline
        left+right-marked (+m+) & 82.3 & 73.58 & 93.2\\
        left-marked (+m) &84.79 & 74.39 & - \\
        \hline
    \end{tabular}
    \label{Tab:trxl-test}
\end{table}

\subsection{Result comparisons for Transformer architectures}
Transformer-XL and BERT both have low perplexity and pseudo-perplexity scores, but both cannot be directly compared as they are calculated quite differently (Eq.\ref{cond}, Eq.\ref{approx}). The dramatically low scores of BERT indicate that per word predicted probability is higher than that of a uni-directional model. Thus the predicted word probability distribution is much sharper when compared to the XL model probability distribution. At this point, we cannot say which model architecture has performed better- BERT or Transformer-XL, despite both of them achieving good low perplexity scores. We would need to experiment with a downstreaming task in-order to fairly compare model performances.

\section{Conclusion}
Recent migration to transformer based architectures in language modeling from LSTM models is justified as 
Transformer-XL obtains strong perplexity results. BERT model also obtains very low pseudo-perplexity scores but it is inequitable to the unidirectional models. Our major contributions in this project, is the use of Transformer-XL architectures for the Finnish language in a sub-word setting, and the formulation of pseudo perplexity for the BERT model. Further comparisons between the transformer architectures can be made by downstreaming it to an ASR task, which will be explored in the future.

\bibliographystyle{ieeetr}
\bibliography{ref_articles.bib}

\begin{thebibliography}{10}

\bibitem{hochreiter1997long}
S.~Hochreiter and J.~Schmidhuber, ``Long short-term memory,'' {\em Neural
  computation}, vol.~9, no.~8, pp.~1735--1780, 1997.

\bibitem{NIPS2017_7181}
A.~Vaswani, N.~Shazeer, N.~Parmar, J.~Uszkoreit, L.~Jones, A.~N. Gomez, L.~u.
  Kaiser, and I.~Polosukhin, ``Attention is all you need,'' in {\em NIPS},
  pp.~5998--6008, Curran Associates, Inc., 2017.

\bibitem{Radford2018ImprovingLU}
A.~Radford, ``Improving language understanding by generative pre-training,''
  2018.

\bibitem{DBLP:journals/corr/abs-1810-04805}
J.~Devlin, M.~Chang, K.~Lee, and K.~Toutanova, ``{BERT:} pre-training of deep
  bidirectional transformers for language understanding,'' {\em CoRR},
  vol.~abs/1810.04805, 2018.

\bibitem{Dai2019TransformerXLAL}
Z.~Dai, Z.~Yang, Y.~Yang, J.~G. Carbonell, Q.~V. Le, and R.~Salakhutdinov,
  ``Transformer-xl: Attentive language models beyond a fixed-length context,''
  in {\em ACL}, 2019.

\bibitem{yang2019xlnet}
Z.~Yang, Z.~Dai, Y.~Yang, J.~Carbonell, R.~Salakhutdinov, and Q.~V. Le,
  ``Xlnet: Generalized autoregressive pretraining for language understanding,''
  2019.
\newblock arxiv:1906.08237.

\bibitem{DBLP:journals/corr/abs-1802-05365}
M.~E. Peters, M.~Neumann, M.~Iyyer, M.~Gardner, C.~Clark, K.~Lee, and
  L.~Zettlemoyer, ``Deep contextualized word representations,'' {\em CoRR},
  vol.~abs/1802.05365, 2018.

\bibitem{DBLP:journals/corr/abs-1801-06146}
J.~Howard and S.~Ruder, ``Fine-tuned language models for text classification,''
  {\em CoRR}, vol.~abs/1801.06146, 2018.

\bibitem{Aaltodoc:http://urn.fi/URN:ISBN:978-952-60-8566-1}
P.~Smit, ``Modern subword-based models for automatic speech recognition,''
  pp.~62 + app. 136, 2019.

\bibitem{inproceedings}
X.~Chen, A.~Ragni, X.~Liu, and M.~Gales, ``Investigating bidirectional
  recurrent neural network language models for speech recognition,''
  pp.~269--273, 08 2017.

\bibitem{10.5555/177910.177914}
P.~Gage, ``A new algorithm for data compression,'' {\em C Users J.}, vol.~12,
  p.~23–38, Feb. 1994.

\bibitem{DBLP:journals/corr/SennrichHB15}
R.~Sennrich, B.~Haddow, and A.~Birch, ``Neural machine translation of rare
  words with subword units,'' {\em CoRR}, vol.~abs/1508.07909, 2015.

\bibitem{DBLP:journals/corr/abs-1907-11692}
Y.~Liu, M.~Ott, N.~Goyal, J.~Du, M.~Joshi, D.~Chen, O.~Levy, M.~Lewis,
  L.~Zettlemoyer, and V.~Stoyanov, ``Roberta: {A} robustly optimized {BERT}
  pretraining approach,'' {\em CoRR}, vol.~abs/1907.11692, 2019.

\bibitem{ftc-korp_en}
{CSC - IT Center for Science}, ``The helsinki korp version of the {F}innish
  text collection, url: http://urn.fi/urn:nbn:fi:lb-2016050207,'' 1998.

\bibitem{smit2014morfessor}
P.~Smit, S.~Virpioja, S.-A. Gr{\"o}nroos, and M.~Kurimo, ``Morfessor 2.0:
  Toolkit for statistical morphological segmentation,'' in {\em Proceedings of
  the Demonstrations at the 14th Conference of the European Chapter of the
  Association for Computational Linguistics}, pp.~21--24, 2014.

\bibitem{10.1145/1187415.1187418}
M.~Creutz and K.~Lagus, ``Unsupervised models for morpheme segmentation and
  morphology learning,'' {\em ACM Trans. Speech Lang. Process.}, vol.~4, Feb.
  2007.

\bibitem{Kingma2014AdamAM}
D.~P. Kingma and J.~Ba, ``Adam: A method for stochastic optimization,'' {\em
  CoRR}, vol.~abs/1412.6980, 2014.

\bibitem{DBLP:journals/corr/ZhuKZSUTF15}
Y.~Zhu, R.~Kiros, R.~S. Zemel, R.~Salakhutdinov, R.~Urtasun, A.~Torralba, and
  S.~Fidler, ``Aligning books and movies: Towards story-like visual
  explanations by watching movies and reading books,'' {\em CoRR},
  vol.~abs/1506.06724, 2015.

\bibitem{DBLP:journals/corr/LoshchilovH16a}
I.~Loshchilov and F.~Hutter, ``{SGDR:} stochastic gradient descent with
  restarts,'' {\em CoRR}, vol.~abs/1608.03983, 2016.

\bibitem{Hmm-dmm}
M.~K. Peter~Smit, Sami~Virpioja, ``Advances in subword-based hmm-dnn speech
  recognition across languages.,'' {\em Submitted to Language Resources and
  Evaluation, 29 November 2018.}

\end{thebibliography}

\end{document}